\documentclass[letterpaper, 10 pt, conference]{ieeeconf}  

\IEEEoverridecommandlockouts                     
\usepackage{url}
\usepackage{algorithm}
\usepackage{algorithmic}
\usepackage{amsmath,amssymb,amsfonts}
\usepackage[font=small,labelfont=bf]{caption}
\usepackage{epsfig,epstopdf,graphics,graphicx,subfig,tabularx}
\usepackage{xcolor}
\usepackage{cite}
\usepackage{caption}
\usepackage{theorem}
\newtheorem{lemma}{Lemma}
\newtheorem{remark}{Remark}
\newtheorem{proposition}{Proposition}

\newtheorem{definition}{Definition}
\newtheorem{theorem}{Theorem} 
\newtheorem{problem}{Problem}

\newcommand{\T}{\mathcal{T}} 

\newcommand{\F}{\mathcal{F}}
\newcommand{\G}{\mathcal{G}}
\newcommand{\A}{\mathcal{A}} 
\newcommand{\B}{\mathcal{B}_{\mathit{all}}} 
\newcommand{\C}{Rew} 
\newcommand{\spec}{\mathbf{\Phi}}
\newcommand{\Prod}{\mathcal{P}} 
\newcommand{\AP}{\Pi} 
\newcommand{\R}{\mathcal{R}} 
\newcommand{\W}{\mathcal{W}}
\renewcommand{\L}{\mathcal{L}} 
\newcommand{\Set}{\mathcal{S}} 
\newcommand{\Nat}{\mathbb{N}} 

\newcommand{\fin}{\mathrm{fin}}

\renewcommand{\P}{\mathcal{P}} 
\newcommand{\Next}{\mathsf{X}}
\newcommand{\Until}{\mathsf{U}}
\newcommand{\Always}{\mathsf{G}}
\newcommand{\Event}{\mathsf{F}}
\newcommand{\true}{\top}
\newcommand{\prop}{\pi}

\newcommand{\ie}{{\it i.e., }}
\newcommand{\eg}{{\it e.g., }}

\newcommand{\BA}{\mathcal B}

\renewcommand{\O}{\mathcal{O}}

\newcommand{\be}{\begin{equation}}
\newcommand{\ee}{\end{equation}}
\newcommand{\ben}{\begin{equation*}}
\newcommand{\een}{\end{equation*}}
\newcommand{\bea}{\begin{eqnarray}}
\newcommand{\eea}{\end{eqnarray}}
\newcommand{\bean}{\begin{eqnarray*}}
\newcommand{\eean}{\end{eqnarray*}}
\newcommand{\ba}{\begin{array}}
\newcommand{\ea}{\end{array}}

\newcommand{\jana}[1]{{ #1}}

\newcommand{\luis}[1]{{#1}}

\definecolor{forestgreen}{rgb}{0,0.6,0.6} 
\definecolor{brown}{rgb}{0.6,0.2,0.2}

\def\myskipabove{\vskip 0cm}
\def\myskip{\vskip 0cm}

\begin{document}

\title{Minimum-\luis{violation} \jana{LTL} Planning with Conflicting Specifications
 \thanks{This work is supported in part by Michigan/AFRL Collaborative Center on Control Sciences, AFOSR grant FA 8650-07-2-3744, the US National Science Foundation, grant CNS-1016213, the National Research Foundation of Singapore, through the Future Urban Mobility SMART IRG, \jana{and grants LH11065 and GAP202/11/0312 at Masaryk University}.}
}

\author{Jana T\r{u}mov\'a \thanks{{J. T\r{u}mov\'a is with Masaryk University. L. I. Reyes Castro, S. Karaman, E. Frazzoli and D. Rus are with Massachusetts Institute of Technology (MIT). This work was initiated while the first author was visiting at MIT and Singapore-MIT Alliance for Research and Technology.}} \and Luis I. Reyes Castro \and Sertac Karaman \and Emilio Frazzoli \and Daniela Rus}

\maketitle
\thispagestyle{empty}
\pagestyle{empty}

\begin{abstract}
We consider the problem of automatic generation of control strategies for robotic vehicles given a set of high-level mission specifications, such as ``Vehicle $x$ must eventually visit a target region and then return to a base," ``Regions $A$ and $B$ must be periodically surveyed," or ``None of the vehicles can enter an unsafe region." We focus on instances when all of the given specifications cannot be reached simultaneously due to their incompatibility and/or environmental constraints. We aim to find the least-violating control strategy while considering different priorities of satisfying different parts of the mission. Formally, we consider the missions given in the form of linear temporal logic formulas, each of which is assigned a reward that is earned when the formula is satisfied. Leveraging ideas from \jana{the} automata-based model checking, we propose an algorithm for finding an optimal control strategy that maximizes the sum of rewards earned if this control strategy is applied. We demonstrate the proposed algorithm on an illustrative~case~study.
\end{abstract}

\section{Introduction}

Control strategy synthesis for robotic systems with high-level, complex, formally-specified goals has recently gained considerable attention in the robotics literature. A diverse set of techniques, including sampling
and cell decomposition of the environment based on triangulations and rectangular partitions
have been used to obtain discrete models of robotic systems; and a variety of temporal logics, including the Computation Time Logic (CTL)~\cite{LaWaAnBe-icra10}, Linear Temporal Logic (LTL)~\cite{KlBe-tac08, WoToMu-cdc09, KrFaPa-tac09, SmTuBeRu-ijrr11, BhKaVa-icra10}, and $\mu$-calculus~\cite{KaFr-cdc09, KaFr-acc12} have been successfully utilized to express complex missions that arise in robotics applications. 
All these references focus on the {\em control synthesis} problem: find a control strategy that satisfies the given specification, if one exists; and report failure otherwise.

The usual execution of many robotic systems, however, involves cases when the mission specification cannot be satisfied as a whole. 
Yet, in most such examples, it is desirable to synthesize a control strategy that fulfills the most important rules, although by (temporarily) violating some of the less important ones.
%
{Consider, for example, an autonomous car navigating in urban traffic. The car must reach its final destination while abiding by the rules of the road, in particular, staying in the right lane and avoiding collision with obstacles. However, for this robot (and a human driver), it is more important not to collide with any other car or pedestrian, than to stay in its own lane. In fact the latter rule is temporarily violated, for instance, when taking over a parked car. }

Another example is from the popular literature. 
Isaac Asimov's ``three laws of robotics'' (see~\cite{asimov}) defines how robots shall interact with humans. According to these laws, a robot may violate any order given by a human operator, if another human life comes in danger. Hence, the latter rule is issued a higher priority than the former one.

Motivated by these examples, 
in this paper, we consider the problem of {\em least-violating control synthesis}, \ie finding a control strategy that satisfies the most important pieces of the mission specification, even if the mission specification can not be fulfilled as a whole. 
The problem can be described as follows. 
Consider a deterministic transition system that models the robot and its environment. The states of the transition system may encode a select set of configurations of the robot (or $n$ robots). Each state of the transition system is labeled with a set of atomic propositions. Examples for atomic propositions include ``The robot is in a safe region," or ``The first robot is in region $A$,'' etc. 
A list of mission specifications, including tasks that need to be fulfilled and rules that must be obeyed, is given in the form of linear temporal logic.  
Each specification in the list is assigned a priority.
Roughly speaking, the least-violating synthesis problem is to find a trace over the transition system that satisfies as many high-priority tasks as possible. 

Our work is related to~\cite{RaKr-cav11, RaKr-icra12}, where the authors study the following problem: given an LTL specification and a model \jana{of a robot} that does not satisfy this specification, decide whether or not the invalidity is limited to the provided model.
{Related literature includes also~\cite{CiRoScTc-vmcai08}, where the authors aim to pinpoint the (un)realizable fragments of the specification to reveal causes of the specification violation.}
]

Other related work includes the recent literature that aims to construct control strategies with minimal changes in the input. 
On one hand, in~\cite{Fa-icra11,KiFaSa-icra12}, the authors aim to find a specification that (i) can be satisfied by the given model, and (ii) is close to the original specification according to a suitable metric. 
On the other hand, in~\cite{Ha-wafr12} the author focuses on finding the least set of constraints (in the model) violating which results in the satisfaction of the specification  \jana{and in~\cite{repair,modelrepair} the authors aim to repair model in the form of a~transition system or a~Markov chain in order to ensure the satisfaction a~given CTL or PCTL formula, respectively.}

Arguably, our work in this paper is closest to the one presented in~\cite{DaFi-fm11}, where the authors consider a transition system with the variables partitioned into control inputs for the car, controllable environment variables and disturbances. The mission specifications are captured as an ordered set of LTL formulas $\Phi = (\phi_1,\ldots,\phi_n)$. The goal is to find the maximal index $1 \leq m \leq n$ and a strategy for the robot ensuring the satisfaction of the subset of formulas $(\phi_1,\ldots,\phi_m)$ regardless of the environmental disturbances. 

Variants of this problem have been addressed also from the perspective of control theory.
For instance, in~\cite{CoYa-tac98}, the authors consider a system modeled as a Markov decision process and a set of specifications given in the form of B\"uchi automata, say $\A_1,\ldots, \A_n$, each of which is assigned a reward, say $rew_1,\ldots, rew_n$. They aim to find a strategy maximizing the total reward gained for the specifications weighted by the respective probabilities with which they are satisfied. The solution builds on translating the problem into a linear programming problem. Unfortunately, the time complexity of their algorithm is exponential in the size of the automata. 

In contrast, our approach takes as input a deterministic transition system, a set of LTL formulas $\phi_1,\ldots, \phi_n$ with rewards $rew_1,\ldots, rew_n$, and we aim to construct a strategy maximizing the total reward gained for the \jana{specifications} that are satisfied. We build the solution on \jana{the} automata-based approach to model checking, which allows us to avoid exponential complexity (in the size of the input automata).

The contribution of this paper can be summarized as follows. We propose an algorithm 
for finding a least-violating trajectory, when the given specification can not be satisfied as a whole. As opposed to a ``brute-force solution'' enumerating all the possible subsets of specifications and attempting to find a strategy for each subset, we build our solution on a single control strategy synthesis procedure, thus substantially reducing the overall computational cost. We demonstrate the proposed approach in an illustrative example. 

The rest of the paper is organized as follows. In Section~\ref{sec:prelims}, we fix some necessary notation and preliminaries. In Section~\ref{sec:pf}, we introduce the problem and outline our approach to its solution. The solution, its correctness and complexity is then discussed in details in Section~\ref{sec:solution}. Section~\ref{sec:example} presents an illustrative case study and we conclude in Section~\ref{sec:conclusion}.

\section{Preliminaries}
\label{sec:prelims}

Given a set $\Set$, let $|\Set|$, $2^\Set$, and $\Set^\omega$ denote the cardinality of $\Set$, the set of all subsets of $S$, and set of all infinite sequences of elements of $\Set$, respectively. A finite and infinite sequence of elements of $\Set$ is called a finite and infinite word over $\Set$, respectively. Given a finite word $w$ and a finite or an infinite word $w'$ over $\Set$, we use $w\cdot w'$ and $w^\omega = w \cdot w \cdot w \ldots$ to denote the word obtained by concatenation of $w$ and $w'$, and by infinitely many repetitions of $w$, respectively.

\subsection{Model and Specification}

\myskipabove
\begin{definition}[Transition System]
A \emph{labeled deterministic transition system} is a tuple $\T=(S,s_{init},\R,\AP,\L)$, where
$S$ is a finite set of states; 
$s_{init} \in S$ is the initial state;
$\R \subseteq S \times S$ is a deterministic transition relation;
$\AP$ is a set of atomic propositions;
$\L: S \rightarrow 2^\AP$ is a labeling function.
\label{def:TS}
\end{definition}
\vspace{-0.18cm}

A \emph{trace} of $\T$ is an infinite sequence of states $\tau = s_0s_1\ldots$ such that $s_0 = s_{init}$ and $(s_i,s_{i+1}) \in \R$, for all $i \geq 0$. A trace $\tau=s_0s_1\ldots $ produces a \emph{word} $w(\tau) = \L(s_0)\L(s_1)\ldots$. 

\myskipabove
\begin{definition}[Formulas of the LTL]
LTL formulas over the set $\AP$ of atomic propositions are constructed inductively according to the following rules:
$$\phi ::= \true \mid \prop \mid \neg \phi \mid \phi \wedge \phi \mid \Next \, \phi \mid \phi\,\Until\,\phi,$$
where $\top$ is a predicate that is always true, $\prop \in \AP$, $\neg$ (negation) and $\wedge$ (conjunction) are standard Boolean operators and $\Next$ (next) and $\Until$ (until) are temporal operators.
\end{definition}
\myskip

LTL formulas are interpreted over infinite words over $2^\AP$, such as those generated by the transition system from Def.~\ref{def:TS}. 
Informally speaking, the word $w = w(0)w(1)\ldots$ \emph{satisfies} the atomic proposition $\prop$  (denoted by $w \models\prop$), if $\prop$ is satisfied in the first position of the word $w$, \emph{i.e.}, if $\prop \in w(0)$. The formula $\Next \, \phi$ states that $\phi$ holds in the following state. The formula $\phi_1\,\Until\,\phi_2$ states that $\phi_2$ is true eventually, and $\phi_1$ is true at least until $\phi_2$ is true. Furthermore, we define formulas $\Event \, \phi \equiv \top \, \Until \, \phi$ and  $\Always\,\phi \equiv \neg(\Event \, \neg \phi)$ that state that $\phi$ holds \emph{eventually} and \emph{always}, respectively.
LTL formulas can express various long term missions, including \emph{surveillance} ($\Always \, \Event \, \phi$, always eventually visit $\phi$), \emph{global absence} ($\Always \, \neg \psi$, globally avoid $\phi$), \emph{reactivity} ($\Always \, \Event \, \phi_1
\Rightarrow \Always \, \Event \, \phi_2$, if $\phi_1$ holds \jana{infinitely often}, then so must $\phi_2$), among many~others. 

The language of all words that satisfy an LTL formula $\phi$ is denoted by $L(\phi)$. With a slight abuse of notation, we extend the satisfaction relation to traces of $\T$, \ie a trace $\tau$ \emph{satisfies} $\phi$ (denoted by $\tau \models \phi$) if and only if the word $w$ produced by $\tau$ satisfies $\phi$. Similarly, a word $w$ and a trace $\tau$ satisfies a set of formulas $\spec$ ($w \models \spec$ and $\tau \models \spec$) if and only if $w \models \phi$ and $\tau \models \phi$, for all $\phi \in \spec$, respectively. 

Given a formula $\phi$, we use $|\phi|$ to denote the \emph{size} of the formula, \ie the number of operators present in $\phi$, and
we use $|\spec|$ to denote $\sum_{\phi \in \spec} |\phi|$.

\myskipabove
\begin{definition}[$\omega$-Automaton]
An $\omega$-automaton is a tuple $\A =  (Q,q_{init},\Sigma,\delta,Acc)$, where
$Q$ is a finite set of states; %
$q_{init}\in Q$ is the initial state; %
$\Sigma$ is an input alphabet; %
$\delta \subseteq Q \times \Sigma \times Q$ is a non-deterministic transition relation; %
$Acc$ is the acceptance condition.
\end{definition}
\myskip

The semantics of $\omega$-automata are defined over infinite input words over $\Sigma$ (such as those generated by transition system from Def.~\ref{def:TS} if $\Sigma = 2^\Pi$). A \emph{run} of the $\omega$-automaton $\A$ \emph{over} an input word $w=w(0)w(1)\ldots$  is a sequence of states
$\rho=q_0q_1\ldots$, such that $q_0  = q_{init}$, and
$(q_i,w(i),q_{i+1}) \in \delta$, for all $i\geq 0$. A \emph{finite run} over a finite word $w_\fin = w(0)\ldots w(l)$ is a finite sequence of states $\rho_\fin = q_0\ldots q_{l+1}$, such that $(q_i,w(i),q_{i+1})) \in \delta$, for all $i \in \{0,\ldots, l\}$.
 
A run $\rho=q_0q_1\ldots$ is \emph{accepting} if it satisfies the acceptance condition $Acc$. For \emph{B\"uchi automata (BA)}, $Acc$ is a set of states $F\subseteq Q$, and $\rho$ is accepting if it intersects $F$ infinitely many times. For \emph{generalized B\"uchi automata (GBA)}, the acceptance condition is a set of sets of states $\F = \{F_1,\ldots, F_m\} \subseteq 2^Q$ and $\rho$ is accepting if it intersects $F_i$ infinitely many times for all $F_i \in \F$. 
A word $w$ is \emph{accepted} by $\A$ if there exists an accepting run over $w$. The \emph{language} of all words accepted by $\A$ is denoted by $L(\A)$.



An $\omega$-automaton is \emph{non-blocking} if for all $q\in Q,\sigma\in \Sigma$ there exists $q'\in Q$, such that $(q,\sigma,q') \in \delta$. For each $\omega$-automaton $\A = (Q,q_{init},\Sigma,\delta,Acc)$ a language equivalent non-blocking $\omega$-automaton can be constructed simply by adding a new state $q_{new}$ to $Q$  and introducing a transition $(q,\sigma,q_{new})$ for all $q \in Q\cup\{q_{new}\},\sigma\in \Sigma$, satisfying the property that $(q,\sigma,q') \not \in \delta$ for all $q'\in Q$.

\myskipabove
\begin{definition}[GBA to BA] A generalized B\"uchi automaton $\G = (Q_\G,q_{init,\G},\Sigma,\delta_\G,\F=\{F_1,\ldots,F_m\})$, can be translated into a B\"uchi automaton 
$\BA = (Q,q_{init},\Sigma, \delta, F)$, such that $L(\BA) = L(\G)$ as follows: $Q = Q_\G \times \{1,\ldots,m\}$; $q_{init} = (q_{init,\G},1)$; $F = F_1 \times \{1\}$; and
$\big((q,j),\sigma,(q',j')\big) \in \delta$ if and only if $(q,\sigma,q') \in \delta_\G$, and
\begin{itemize}
\item $q \not \in F_j$ and $j' = j$, or
\item $q \in F_j$ and $j'=(j \mod m) +1$.
\end{itemize}
\label{def:gba}
\end{definition}
\begin{definition}[Automata Intersection]
Given $n$ B\"uchi automata $\BA_1,\ldots,\BA_k$ where $\BA_i = (Q_i, q_{init,i}, \Sigma, \delta_i, F_i)$ for all $1\leq i \leq n$, a B\"uchi automaton $\BA = (Q,q_{init},\Sigma,\delta,F)$, such that $L(\BA) = L(\BA_1) \cap\ldots \cap L(\BA_n)$ can be built as follows:
$Q=Q_1\times \ldots \times Q_n \times \{1,\ldots,n\}$; 
$q_{init}= (q_{init,1}, \ldots ,q_{init,n}, 1)$; 
$F = F_1 \times Q_2 \times \ldots \times Q_n \times \{1\}$; and
$\big((q_1,\ldots,q_n,j),\sigma, (q_1',\ldots, q_n',j')\big) \in \delta$ if and only if $(q_i,\sigma,q_i') \in \delta_i$, for all $i \in \{1,\ldots, n\}$, and
\begin{itemize}
\item $q_j \not \in F_j$ and $j' = j$, or
\item $q_j \in F_j$ and $j' = (j \mod n) +1$.
\end{itemize}
\label{def:interesction}
\end{definition}
\myskip

Intuitively, the set of states of $\BA$ can be viewed as $n$ copies (layers) of the Cartesian product of the sets of states $Q_1\times \ldots \times Q_n$. 

\smallskip

Any LTL formula $\phi$ over $\AP$ can be translated into a B\"uchi automaton $\BA_\phi$ with alphabet $2^\AP$, such that $L(\phi) = L(\BA_\phi)$. A number of standard translation algorithms (see, \eg \cite{GePeVaWo-96,GaOd-cav01}) rely on a three-step procedure: First, the formula is normalized, second, it is translated into a generalized B\"uchi automaton and third, the obtained GBA is finally translated into a language-equivalent B\"uchi automaton (see Def.~\ref{def:gba}).

\smallskip

A weighted $\omega$-automaton  $\A =  (Q,q_{init},\Sigma,\delta,Acc, \W)$ is an $\omega$-automaton, where  $Q,q_{init},\Sigma,\delta,Acc$ are defined in the usual way, and $\W: \delta \rightarrow \Nat$ is a function assigning a weight to each transition.

Let $\rho = q_0q_1 \ldots$ and $\rho_\fin = q_0\ldots q_{l+1}$ be an accepting run over \jana{$w = w(0)w(1)\ldots$} and a finite run over \jana{$w_\fin = w(0) \ldots w(l)$} of  a weighted B\"uchi automaton $\BA$, respectively. We use $\mathsf{Fra}\mathsf{g} (\rho) = \{q_i\ldots q_k \mid q_i, q_k \in F \text{ and } q_j \not \in F \text{ for all } i<j<k\}$ and $\mathsf{Frag}(\rho_\mathrm{fin}) =  \{ q_{i}\ldots q_{k} \mid \  q_{i}, q_{k} \in \jana{F}, 0\leq i \leq k \leq l \text{ and } \nonumber q_j \not \in \jana{F}, \text{ for all } i<j<k\}$
 to denote the set of all finite fragments of $\rho$ and $\rho_\fin$ that begin and end in an accepting state and do not contain any other accepting state. Note that each accepting run $\rho$ and each finite run $\rho_\fin$ corresponds to a unique sequence of fragments. With a slight abuse of notation, we use 
 \begin{align*} 
 \W(q_i\ldots q_k) =  \sum_{j=i}^{k-1} \W((q_j,\jana{w(j)},q_{j+1}))
 \end{align*}
to denote the sum of the weights between the states of fragment $q_i\ldots q_k$ of a run $\rho$ over \jana{$w$} (or a finite run $\rho_\fin$ over \jana{$w_\fin$}).

\subsection{Automata-Based Model Checking and Strategy Synthesis} 

Given a transition system $\T$ and a B\"uchi automaton $\BA$, 
the \emph{model checking problem} is to prove or disprove that all traces of $\T$ satisfy $\BA$, whereas the \emph{control strategy synthesis problem} is to find a trace of $\T$ that satisfies $\BA$.
Both of these problems can be addressed by constructing a product automaton $\P$ that captures all the behaviors of $\T$ satisfying $\BA$ and searching for an accepting run of $\P$.
\myskipabove
\begin{definition}[Product Automaton]
A product automaton of a transition system $\T=(S,s_{init},\R,\AP,\L)$ and a BA $\BA =  (Q,q_{init},\Sigma,\delta,F)$ is a B\"uchi automaton $\P =\T \otimes \BA=  (Q_\P,q_{init,\P},\delta_\P,F_\P)$, where $Q_\P = S \times Q$; $q_{init,\P} = (s_{init},q_{init})$;  $F_\P = S \times F$; and $((s,q),(s',q')) \in \delta_\P$ if 
\begin{itemize}
\item $(s,s') \in \R$ and $(q,\L(s),q')\in \delta$
\end{itemize}
If $\BA= (Q,q_{init},\Sigma,\delta,F, \W)$ is a weighted B\"uchi automaton, $\P$ is also weighted: $\P = (Q_\P,q_{init,\P},\delta_\P,F_\P,\W_\P)$, where $\W_\P\big(((s,q),(s',q'))\big) = \W\big((q,\L(s),q')\big)$, for all $((s,q),(s',q'))\in \delta_\P$. 
\label{def:product}
\end{definition}
\myskip
The product automaton has a trivial alphabet, which is therefore omitted. An accepting run $\rho$ of the product automaton projects onto a trace $\tau$ of $\T$ (denoted by $\tau = \alpha(\rho)$) that satisfies the property captured by the B\"uchi automaton $\BA$. Vice versa, any trace of $\T$ satisfying the property corresponds to an accepting run of the product automaton. Furthermore, if there exists an accepting run $\rho_\P$ of $\P$, then there exists an accepting run $\rho_\P'$ of $\P$ in a \emph{prefix-suffix structure}, \ie $\rho_\P' = \rho_\mathrm{pref}\cdot(\rho_\mathrm{suf})^\omega$ for some finite sequences $\rho_\mathrm{pref}$ and $\rho_\mathrm{suf}$ of states of $\P$, such that the first state of $\rho_\mathrm{suf}$ is an accepting state from $F_\P$.

The (weighted) product automaton can be viewed as a (weighted) graph $(V,E)$ with the set of vertices $V$ equal to the set of states $Q_P$ and the set of edges $E$ (and their weights) given by the transition function $\delta_\P$ (and the weight function $\W_\P$) in the expected way. 
A \emph{simple path} in $\P$ is a sequence of states $p_i\ldots p_l$ such that $(p_j,p_{j+1})\in \delta_\P$, for all $i \leq j< l$, and  $p_j = p_{j'} \Rightarrow j=j'$, for all $i\leq j,j' \leq l$. A \emph{cycle} is a sequence of states $p_i\ldots p_lp_{l+1}$, where $p_i \ldots p_l$ is a simple path and $p_{l+1}=p_i$. A state $p'$ \emph{reachable} from $p$ if there is a simple path from $p$ to $p'$.

\myskipabove
\begin{definition}[Maximal simple distance] 
\emph{The maximal simple distance} from $p_f \in F_\P$ to $p$ in a weighted product automaton $\P$ is the maximal sum of edge weights on a simple path $p_i \ldots p_l$ from $p_i=p_f$ to $p_l=p$, such that $p_j\not \in F_\P$, for all $i<j<l$.
\label{def:maxdist}
\end{definition}
\myskip
Efficient graph search algorithms can be used for finding a prefix $\rho_\mathrm{pref}$ (a simple path from the initial state to an accepting state in the product graph) followed by a periodically repeated suffix $\rho_\mathrm{suf}$ (a cycle in the product graph containing an accepting state) of an accepting run $\rho = \rho_\mathrm{pref} \cdot (\rho_\mathrm{suf})^\omega$ (a lasso-shaped path in the product graph).
One of the
standard algorithms to do so is \emph{nested depth-first search (DFS)}~\cite{principles}, successfully implemented, for instance, in the pioneer model checker SPIN~\cite{spin}. The (worst-case) running time complexity of this algorithm is linear in time and space with respect to the size (the number of states and transitions) of the product automaton $\P$.

\section{Problem Formulation and Approach}
\label{sec:pf}

Let us consider a robot moving in a partitioned environment with its motion capabilities modeled as a labeled transition system $\T = (S,s_{init},\R,\AP,\L)$ from Def.~\ref{def:TS}. Each region of the environment is modeled as a state of the transition system and the robot's ability to move between two regions is represented as transition between the corresponding states. In case several controlled robots are placed in the environment, the states of the transition systems encode positions of all the robots in the environmental regions \ie for $k$ robots, a state corresponds to an $k$-tuple of regions, where the $i$-th element of the tuple is the region in which the $i$-th robot is placed. The transitions between the states reflect the simultaneous motion capabilities of all the robots. The labeling function $\L$ maps each state of the transition system to a subset of atomic propositions from $\AP$ that hold true in this state, such as ``Vehicle $x$ is in a safe region.".

There is a set of high-level missions to be accomplished by the robotic system expressed as a set of LTL formulas $\spec = \{\phi_1,\ldots,\phi_n\}$ over $\AP$ with priorities of their satisfaction determined by a reward function $rew: \spec \to \Nat$. The value $rew(\phi_i)$ represents the reward that is gained if specification $\phi_i$ is accomplished. Without loss of generality, from now on, we assume that $rew(\phi_i) \geq rew(\phi_j)$, for all $1\leq i \leq j \leq n$.


Given a trace $\tau$ of the transition system $\T$, we define \emph{trace reward} as the sum of the rewards of all formulas from $\spec$ that are satisfied on this run.
\myskipabove
\begin{definition}[Trace Reward]
\emph{Reward of a trace} $\tau$ of $\T$ is 
\begin{align}
Rew(\tau) = \sum_{\{\phi_i \mid \tau \models \phi_i\}} rew(\phi_i).
\label{eq:reward}
\end{align}
\end{definition}

We are now ready to formally state our problem of finding "the best" trace of $\T$, \ie "the least violating" motion of the robot (or the robots) in the environment with respect to the given set of mission specifications.
\myskipabove
\begin{problem}
 \emph{Given} \begin{itemize}
 \item a transition system $\T=(S,s_{init},\R,\AP,\L)$;
 \item a set of LTL formulas $\spec = \{\phi_1, \ldots, \phi_n\}$ over $\AP$; and
 \item  a reward function $rew: \spec \rightarrow \Nat$, 
 \end{itemize}
 \emph{find} a trace $\tau$ of $\T$ that maximizes $Rew(\tau)$ from Eq.~\ref{eq:reward}. 
\label{prob:reward}
\end{problem}
\begin{remark}  
Note, that if $rew(\phi_i) = 2^{n-i}$, for each formula $\phi_i \in \spec$, then the set $\spec$ is in fact ordered according to the standard lexicographic ordering. In other words, it is always more important to satisfy $\phi_i$ than $\phi_{i+1} \wedge \ldots \wedge \phi_n$.
\end{remark}

A straightforward solution to Prob.~\ref{prob:reward} is to consider all the possible subsets $\spec_I = \{\phi_i \mid i \in I\}$, $I \subseteq \{1,\ldots,n\}$ of formulas from $\spec$ and to find a trace $\tau_I$ of $\T$ satisfying $\spec_I$ if such a trace exists. The search can be done using one of the known model-checking algorithms (\eg the automata-based algorithm from Sec.~\ref{sec:prelims}). A trace $\tau_I$ maximizing $Rew(\tau_I)$ among the found ones maps to the desired robot path. However, this brute-force solution is not efficient as it requires up to $2^n$ model-checking procedure runs in the worst case. 

In this paper, we suggest a method to alleviate the high computational demand of this straightforward solution. The main idea builds on \jana{the} automata-based approach to model-checking. We construct a single weighted B\"uchi automaton $\B$ for formula $\bigwedge_{\phi_i \in \spec} \phi_i$ and capture the rewards of the LTL formulas through its weights. Then, a weighted product automaton $\P = \T \otimes \B$ is built and an optimal accepting run of $\P$ is sought using a modification of the nested-DFS algorithm, with the computational complexity only slightly worse in comparison to the original nested-DFS. Roughly speaking, instead of up to $2^n$ model-checking procedure runs, we perform only a single execution of an altered model-checking algorithm.

\section{Problem Solution}
\label{sec:solution}

This section introduces our solution to Prob.~\ref{prob:reward} in detail. First, we present the construction of the weighted B\"uchi automaton $\B$ and the weighted product automaton $\P$. Second, the modified nested-DFS is given. Third, we discuss the solution correctness, completeness and complexity.

\subsection{Construction of the Weighted Automata}

Consider the set of mission specifications $\spec = \{\phi_1,\ldots, \phi_n\}$ that are translated (\eg using the algorithm from~\cite{GaOd-cav01}) into generalized B\"uchi automata 
\begin{align*} 
& \G_{\phi_1} =  (Q_1,q_{init,1},\Sigma,\delta_1,\F_1=\{F_1^1,\ldots,F_1^{m_1}\}) ,\ldots  \\ 
\ldots , \ & \G_{\phi_n} = (Q_n,q_{init,n},\Sigma,\delta_n,\F_n=\{F_n^1,\ldots,F^{m_n}_n\}), 
\end{align*}
respectively. Without loss of generality, we assume that $\G_{\phi_1},\ldots,\G_{\phi_n}$ are all non-blocking. We build the weighted B\"uchi automaton $\B$ leveraging ideas from translation of generalized B\"uchi automata to B\"uchi automata (Def.~\ref{def:gba}) and from construction of a B\"uchi automaton for language intersection of several B\"uchi automata (Def.~\ref{def:interesction}). 

\myskipabove
\begin{definition}[Weighted B\"uchi automaton]
A weighted 
B\"uchi automaton $\B = (Q,q_{init},\Sigma,\delta,F,\W)$ is defined as follows:
\begin{itemize}\itemsep0ex
\item \jana{$Q=Q_1\times \ldots \times Q_n \times \\ (\{(j,l) \mid 1 \leq j \leq n, 1 \leq l \leq m_i \} \cup \{(0,0)\})$;}
\item $q_{init}= (q_{init,1}, \ldots, q_{init,n},(0,0))$;
\item $t = \big((q_1,\ldots,q_n,(j,l)),\sigma, (q_1',\ldots, q_n',(j',l'))\big) \in \delta$ if $(q_i,\sigma,q_i') \in \delta_i$, for all $i \in \{1,\ldots, n\}$, and
\begin{enumerate}
\item $(j,l)=(0,0)$ and 
\begin{enumerate}
\item $(j',l') = (0,0)$. Then $\W\big(t\big) = 0$.
\item $j' > 0, l'=1$. Then $\W\big(t\big) = rew(\phi_{j'})$.
\end{enumerate}
\item $j \neq 0$ and
\begin{enumerate}
\item $(j',l') =  (j,l) $ and $q_j \not \in F_j^l$. Then $\W\big(t\big) = 0$.
\item $ l\neq m_j, (j',l')=(j, l+1)$ and $q_j \in F_j^l$. Then $\W\big(t\big) = 0$.
\item $l= m_j, j < n$, $j \leq j'$, $l'=1$ and $q_j \in F_j^l$. Then $\W\big(t\big) = rew(\phi_{j'})$.
\item $l = m_j, (j',l')= (0,0)$, and $q_j \in F_j^l $. Then $\W\big(t\big) = 0.$
\end{enumerate}
\end{enumerate}
\item $F = Q_1 \times Q_2 \times \ldots \times Q_n \times \{(0,0)\}$.
\end{itemize}
\label{def:weighted}
\end{definition}
\myskip


\jana{Loosely speaking, the set of states of the automaton $\B$ can be viewed as \emph{layers}, where the $j$-th layer consists of $m_j$ \emph{components}, for all $1 \leq j \leq n$. Each component then involves a copy of each element from the Cartesian product $Q_1\times\ldots \times Q_n$. Within the $j$-th layer, the $l$-th component is connected to the $(l+1)$-th component through transitions leading from $F_j^l$. The $j$-th layer is connected to the $j'$-th through transitions leading from $F_j^{m_j}$, for all $j+1 \leq j' \leq n$. These transitions are labeled with the reward $rew(\phi_{j'})$. Besides that, the layer $0$ consist only one component $(0,0)$, whose states are all and the only ones accepting. From this component, transition leads to the first component of each layer, and dually, from the last component of each layer, transitions lead to this component. 

Note that the automaton $\B$ accepts all words satisfying specifications $\bigwedge_{\phi_i \in \spec_I} \phi_i$, for all $\spec_I \subseteq \spec$. The weights associated with transitions connecting the layers determine the ``quality'' of a particular run, \ie they capture which formulas are satisfied by this run. Particularly, if an accepting run enters the $j$-th layer infinitely many times, then it intersects all $F_j^l \in \F_j$ infinitely many times and thus the satisfaction of $\phi_j$ is guaranteed. Furthermore, such a run contains infinitely many transitions weighted with $rew(\phi_j)$.
}

Formally, the purpose of the weights of $\B$ is summarized as follows. \jana{Let us denote the component of a state as $component(q_1,\ldots,q_n,(j,l))=(j,l)$.}
\myskipabove
\begin{definition}[Run Reward] The {reward}
of a run $\rho$ of $\B$ is
\begin{align*}
\C(\rho) = \max \big\{C \mid \ & C = \W(q_i\ldots q_{l}) \text{ for infinitely many} \\ 
& \text{ fragments } q_i\ldots q_l \in \mathsf{Frag}(\rho)\big\}.
\end{align*}
\end{definition}
\myskip

Intuitively, a run $\rho$ can be split into a sequence of fragments that is associated with a respective sequence of fragment weights. The run reward is equal to the maximal weight that appears in the sequence of fragment weights infinitely many times.
\myskipabove
\begin{lemma}
Consider a word  $w = w(0)w(1) \ldots $, where $w \models \spec_I$ and $w \not \models \phi$, for all $\phi \not \in \spec_I$. There exists an accepting run $\rho = q_0q_1\ldots$ of $\B$ over $w$, such that $\C(\rho) = \sum_{\phi_i \in \spec_I} rew(\phi_i). $ 
Furthermore, for each accepting run $\rho'=q_0'q_1'\ldots$ of $\B$ over $w$ it holds, that $\C(\rho') \leq \sum_{\phi_i \in \spec_I} rew(\phi_i).$
\label{lemma:1}
\end{lemma}
\myskip
\begin{proof}
If $w \models \spec_I$ then there is an accepting run $\rho_{i} = q_0q_1\ldots$, for all $\phi_i \in \spec_I$. Let $I=\{i_1,\ldots,i_j\}$. According to the construction of $\B$ (Def.~\ref{def:weighted}), there exists a run $\rho = p_0p_1\ldots$ of $\B$, such that each fragment $p_{k}\ldots p_{k'} \in \mathsf{Frag}(\rho)$ satisfies the following.
\begin{align*}
\jana{component}(p_{l_1}) = & \ (0,0) \\
\jana{component}(p_{(l_1+1)}) = &\ldots = \jana{component}(p_{l_2}) =(i_1,1)\\ 
& \ldots \\
\jana{component}(p_{l_3}) = &\ldots = \jana{component}(p_{l_4}) =(i_1,|\F_{i_1}|)\\ 
\jana{component}(p_{(l_4+1)}) = & \ldots = \jana{component}(p_{l_5}) = (i_2,1) \\
& \ldots \\
\jana{component}(p_{l_6}) = &\ldots = \jana{component}(p_{l_7}) = (i_2, |\F_{i_2}|)\\
& \ldots \\
\jana{component}(p_{(l_8)}) = & \ldots = \jana{component}(p_{l_{9}}) = (i_j,1) \\
& \ldots \\
\jana{component}(p_{l_{10}}) = &\ldots = \jana{component}(p_{l_{11}}) = (i_j, |\F_{i_j}|)\\
\jana{component}(p_{l_{({11}+1)}}) = & \ (0,0)
\end{align*}
where $p_{l_1} = p_k$, $p_{l_{({11}+1)}} = p_{k'}$. The total weight of such a fragment and hence also the reward of $\rho$ is equal to $\sum_{\phi_i \in \spec_I} \phi_i$ directly from the construction of $\B$.

On the other hand, assume that there exists a run $\rho'=q_0'q_1'\ldots$ of $\B$ over $w$ such that $\C(\rho') > \sum_{\phi_i \in \spec_I} rew(\phi_i).$ From the construction of the automaton $\B$, this means that there exist infinitely many fragments $p_{k}\ldots p_{k'} \in \mathsf{Frag}(\rho')$ with their weight larger than $\sum_{\phi_i \in \spec_I} rew(\phi_i)$. Therefore, there exist $\phi_l \not \in \spec_I$, and states $p_1',\ldots,p_{|\F_l|}'$ of $\B$, such that $\jana{component}(p_j') = (l,j)$, for all $j \in \{1,\ldots,|\F_l|\}$. Thus, the run $\rho'$ can be projected to an accepting run of $\BA_l$ over $w$, which is in contradiction with our assumption that $w \not \models \phi_l$ for all $\phi_l \not \in \spec_I$. 
\end{proof}
\medskip
The second step of our algorithm is the construction of a product automaton $\P =\T \otimes \B=  (Q_\P,p_{init},\delta_\P,F_\P,\W_\P)$ (see Def.~\ref{def:product}). Based on Lemma~\ref{lemma:1}, the product automaton satisfies the following:
\myskipabove
\begin{lemma}
Let $\tau$ be a trace of $\T$. Then, there exists a run $\rho_\Prod$ of $\Prod$ \jana{with} $\tau = \alpha(\rho_\Prod)$ such that the reward $Rew(\tau) = \C(\rho_\Prod)$. Moreover, $Rew(\tau) \geq \C(\rho_\Prod')$ for all $\rho_\Prod'$ with $\alpha(\rho_\Prod') = \tau$.
\label{lemma:traces}
\end{lemma} 
\myskip
\begin{proof}
The proof follows directly from Lemma~\ref{lemma:1} and the fact that for each trace $\tau$ that produces a word $w=w(0)w(1)\ldots$ accepted by a run $\rho=q_0q_1\ldots$ of $\B$, there exists an accepting run $\rho_\Prod=p_0p_1\ldots$ in $\Prod$, such that $\W((p_i,p_{i+1}))=\W((q_i,w(i),q_{i+1}))$  and $q_i \in F \iff p_i \in F_\Prod$, for all $i\geq 0$.
\end{proof}
\myskipabove
\begin{lemma}
For each run $\rho_\P$ there exists a run $\rho_\P'$ in {prefix-suffix structure}, such that $\C(\rho_\P)=\C(\rho_\P')$.
\label{lemma:pf-sf1}
\end{lemma}
\myskip
\begin{proof}
Because $\rho_\P = p_0p_1\ldots$ is infinite, there exist a state $p\in F_\Prod$ that appears on $\rho_\P$ infinitely many times and there exist a fragment $p\ldots p'$ starting in $p$ such that $\W(p\ldots p') = 
\C(\rho_\P)$. Because $p$ occurs on $\rho_\P$ infinitely many times, $p$ is reachable from $p'$. Therefore, run $\rho_\P$ is a sequence of states $\rho_\P = p_0p_1 \ldots p \ldots p' \ldots p \ldots$. Let $\rho_\P' = p_0p_1\ldots (p\ldots p' \ldots p)^\omega$. Run $\rho_\P'$ is in prefix-suffix structure, it is accepting and $\C(\rho_\P') = \C_(\rho_\P)$. 
\end{proof}
\bigskip

The \jana{three} lemmas above provide us with guidance on computing the trace of $\T$ with the maximal reward: it is enough to compute a run of $\P$, in the prefix-suffix structure, that maximizes $\C(\rho_\Prod)$ and project this run into a trace of~$\T$. This is stated in the following proposition.
\myskipabove
\begin{proposition}
Let $\tau = s_0s_1\ldots$ be a trace of $\T$, such that $\tau \models \spec_I$ and $\tau \not \models \phi$, for all $\phi \not \in \spec_I$. Then, there exists an accepting run $\rho_\P=(s_0,q_0)(s_1,q_1)\ldots$ in $\P$ such that 
\begin{itemize}
\item[(i)] $\rho_\P$ is in prefix-suffix structure  and 
\item[(ii)] $\C(\rho_\P) = \sum_{\phi_i \in \spec_I} \phi_i $. 
\end{itemize}
\label{prop:pf-sf}
\end{proposition}

The remaining task is to find a run $\rho_\P$ satisfying the condition (i) of Proposition~\ref{prop:pf-sf} and maximizing $\C(\rho_\P)$.
The problem thus reduces to searching for a reachable cycle $c$ (a repeated run suffix) in $\P$ beginning (and thus also ending) in an accepting state that maximizes the value 
\begin{align}
\C(c)=\max_{p_i\ldots p_l \in \mathsf{Frag}(c)} \W(p_i \ldots p_{l})
\label{eq:max}
\end{align}
among all such cycles. 
The following lemma helps narrow down the search even further, showing that it is enough to search for a particular type of cycle.
\myskipabove
\begin{lemma}
Given a cycle $c$ in $\P$ and a fragment  $p_i \ldots p_l \in \mathsf{Frag}(c)$, there exists a simple path $p_i \ldots p_l$, such that $\W(p_i\ldots p_{l}) = 0$. 
\label{lemma:cycle}
\end{lemma}
\myskip
\begin{proof}
From the construction of the automaton $\B$, it follows that if there is a simple path from $(q_1,\ldots,q_n,(0,0)) \in Q$ to $(q_1',\ldots,q_n',(i,j)) \in Q$ in the automaton $\B$, then there exists a simple path from $(q_1,\ldots,q_n,(0,0)) $ to $(q_1',\ldots,q_n',(0,0))$ that contains only states $\mathbf{q} \in Q$, such that $\jana{component}(\mathbf{q})=(0,0)$.
Thus, if there is a simple path from $(s,q_1,\ldots,q_n,(0,0)) \in Q_\P$ to $(s',q_1',\ldots,q_n',(i,j)) \in Q_\P$ in the product automaton $\P$, then there exists also a simple path from $(s,q_1,\ldots,q_n,(0,0)) $ to $(s',q_1',\ldots,q_n',(0,0))$ that contains only states $p\in Q_\P$, such that $\jana{component}(p)=(0,0)$ The reward of such a simple path is 0.
\end{proof}
\smallskip
Thanks to Lemma~\ref{lemma:cycle}, it is enough to search for a cycle $c$ maximizing~Eq.~\ref{eq:max}, such that  $\W(p_i \ldots  p_{l}) = 1$, for all fragments $p_i \ldots p_l  \in \mathsf{Frag}(c)$, but one. Hence, without loss of generality, we can consider only cycles $c= p_i\ldots p_l p_{l+1} \ldots p_i$ such that \jana{$\W(p_i \ldots p_{l}) \neq 0$} only for the first fragment $p_i\ldots p_l$ of the cycle. Such a cycle can be found by adapting standard nested depth-first search algorithm as we will show in the following section. 

\myskipabove
\begin{proposition}
A maximal-reward trace of $\T$ can be obtained as a projection $\alpha(p_0\ldots p_l)\cdot (\alpha(c))^\omega$ of a path $p_0\ldots p_l$ and a cycle $c = p_{l+1}\ldots p_ip_{i+1} \ldots p_{l+1}$, such that
\begin{itemize}
\item $p_0 = q_{init,\P}$, $(p_l,p_{l+1}) \in \delta$, $p_{l+1} \in F_\P$,
\item $\W(p_{l+1} \ldots p_i)$ for the first fragment $p_{l+1} \ldots p_i \in \mathsf{Frag}(c)$ of the cycle is maximized, and
\item \jana{$\W(p_j \ldots p_k) = 0$}, for all fragments $p_j\ldots p_k \in \mathsf{Frag}(p_{i+1}\ldots p_{l+1})$.
\end{itemize}
\label{prop:projection}
\end{proposition}

\subsection{Weighted Nested Depth-First Search}

This section aims at search for a path $p_0\ldots p_l$ followed by a cycle $p_{l+1}\ldots p_{l+1}$ satisfying conditions of Prop.~\ref{prop:projection}. The solution is summarized in Alg.~\ref{alg:ndfs} to Alg.~\ref{alg:cycle}. \jana{The external functions used in the algorithms are summarized and explained in Table~\ref{tab:functions}.}

\jana{First, let us focus on a solution to the following sub-problem: Given an accepting state $p_f \in F_\Prod$, find a cycle $c=p_f \ldots p_ip_{i+1}\ldots p_f$ that maximizes value $\W(p_f\ldots p_i)$  in Eq.~\ref{eq:max} for the first fragment $p_f \ldots p_i \in \mathsf{Frag}(c)$ among all cycles that begin and end in $p_f$.
A modification of breath-first graph search algorithm as described in Alg.~\ref{alg:cycle} can be used to do so 
in $\mathcal{O}(|\P|)$ time and space
thanks to the fact that the individual layers connected through non-zero weighted transitions form a directed acyclic graph.
Intuitively, the algorithm systematically searches the graph $\P$ and maintains for each state $p$ the approximation of the maximal simple distance (Def.~\ref{def:maxdist})
from $p_f$ to $p$. The correctness of the algorithm relies on the fact, that when $p$ is processed on line~\ref{line:2} of the procedure $\mathsf{propagate}$ (Alg.~\ref{alg:propagate}), the value of $p.dist$ is set to the actual maximal simple distance from $p_f$ to $p$. When all states that are reachable from state $p_f$ are visited in Alg.~\ref{alg:propagate}, the second phase (lines~\ref{line:while}-\ref{line:endwhile}) of Alg.~\ref{alg:cycle} is executed to check whether $p_f$ is also reachable from $p$, considering $p$ one by one in descending order of their $p.dist$.

}

\jana{
Second, the cycle satisfying conditions of Prop.~\ref{prop:projection} can be found by running Alg.~\ref{alg:cycle} from each $p_f \in F$ reachable from the initial state, potentially traversing the whole graph $|F|$-times. However, leveraging ideas from nested DFS algorithm, the complexity can be reduced. The idea is to run Alg.~\ref{alg:cycle} from states in $F_\P$ in particular order that ensures the states visited during previous executions of Alg.~\ref{alg:cycle} do not need to be visited again. In particular, in the standard nested DFS it holds that if a cycle is being sought from a state $p_f'$ (so-called inner-search) that is reachable from $p_f$ and the search is unsuccessful, then later, when a cycle is sought from $p_f$, the states visited in the inner-search from $p_f'$ do not have to be considered again. Based on this idea, we formulate the following lemma that explains the correctness of our approach.
}
\myskipabove
\begin{lemma}
Let $p_f' \in F_\P$ be reachable from $p_f \in F_\P$ and $p \in Q_\P$ be reachable from both $p_f$ and $p_f'$. If there exists a cycle $c$ from state $p_f \in F_\P$ containing state $p$, then there exists a cycle $c'$ from $p_f' \in F_\P$ with reward $\C(c') \geq \C(c)$.
\label{lemma:reachability}
\end{lemma}
\myskip
\begin{proof}
Because $p_f$ is reachable from $p$, $p$ is reachable from $p_f'$, and $p_f'$ is reachable from $p_f$, then $p_f$ is reachable from $p_f'$. Therefore, there exists a cycle $c' = p_f'\ldots p_f \ldots p \ldots p_f \ldots p_f'$, where $p_f\ldots p \ldots p_f = c$. Clearly $\C(c') \geq \C(c)$.
\end{proof}
\smallskip

\begin{table}[!h]
\begin{tabular}{ll}
$\mathsf{find\_arbitrary\_trace(\T)}$ & returns and arbitrary trace of TS $\T$ \\
$\mathsf{find\_path}(\P,p,p_f)$ & returns a path from $p$ to $p_f$ in $\P$ \\
$\mathsf{successors}(p)$ & returns the immediate successors of $p$ in $\P$\\
$stack.\mathsf{push}(p)$ & inserts $p$ on the top of $stack$ \\
$stack.\mathsf{top}()$ & reads from the top of $stack$ \\
$stack.\mathsf{top\_and\_pop}()$ & destructively reads from the top of $stack$ \\
$stack.\mathsf{pop}()$ & removes element from the top of $stack$ \\
$\mathsf{reverse}(stack)$ & returns the elements of $stack$ in the reversed \\&  order
\end{tabular}
\caption{\jana{List of functions used in Alg.~\ref{alg:ndfs}--\ref{alg:propagate}}}
\label{tab:functions}
\end{table}

\begin{algorithm}[!h]
\small
\caption{\small $\mathsf{weighted\_nested\_DFS}(\P)$}
\begin{algorithmic}[1]
\INPUT{product automaton $\P$}
\OUTPUT{solution to Prob.~\ref{prob:reward}}
\STATE $weight\_max = 0; prefix\_max = \epsilon; cycle\_max = \epsilon$
\STATE $stack\_outer= \mathrm{empty}; visited\_outer = \varnothing$ 
\STATE $visited\_inner = \varnothing; visited\_ps = \varnothing$
\FORALL {$p \in Q_\P$}
\STATE $p.dist = 0; p.pred = \bot$
\ENDFOR
\STATE $run:=\mathsf{DFS}(\P,p_{init})$
\IF {$run\neq \epsilon$}
\RETURN $trace:= \alpha(run)$
\ELSE
\RETURN $\mathsf{find\_arbitrary\_trace(\T)}$
\ENDIF
\end{algorithmic}
\label{alg:ndfs}
\end{algorithm}

\begin{algorithm}[!h]
\small
\caption{\small $\mathsf{DFS}(\P,p)$}
\begin{algorithmic}[1]
\jana{\INPUT{product automaton $\P$, state $p$}
\OUTPUT{run of $\P$ satisfying conditions of Prop.~\ref{prop:projection}.}}
\STATE $stack\_outer.\mathsf{push}(p); visited\_outer := visited\_outer \cup \{p\}$
\REPEAT
\STATE $p' := stack\_outer.\mathsf{top}()$
\IF {$\mathsf{successors}(p') \setminus visited\_outer \neq \varnothing$}
\STATE pick $p'' \in \mathsf{successors}(p') \setminus visited\_outer$
\STATE $stack\_outer.\mathsf{push}(p'')$
\STATE $visited\_outer :=~visited\_outer \cup \{p''\}$
\ELSE
\STATE $stack\_outer.\mathsf{pop}()$
\IF {$p' \in F_\P$}
\STATE $p'.dist := 0$; $p'.pred=\bot$
\STATE $cycle:= \mathsf{longest\_cycle\_search}(\P,p')$
\IF {$p'.dist > weight\_max$}
\STATE $weight\_max:=p'.dist; cycle\_max:=cycle$
\STATE $prefix\_max:=\mathsf{reverse}(stack\_outer)$
\ENDIF
\ENDIF
\ENDIF
\UNTIL{$(stack\_outer = \mathrm{empty} \vee weight\_max=n)$}
\RETURN {$prefix\_max \cdot (cycle\_max)^\omega $}
\end{algorithmic}
\label{alg:dfs}
\end{algorithm}

\begin{algorithm}[h!]
\caption{\small $\mathsf{longest\_cycle\_search}(\P,p_f)$}
\begin{algorithmic}[1]
\small
\INPUT{product automaton $\P$, accepting state $p_f\in F_\P$}
\OUTPUT{cycle $(p_f,\ldots,p_f)$ maximizing Eq.~\ref{eq:max} (if one exists)}
\STATE $queue\_curr := (p_f)$
\FORALL {$1 \leq i \leq n$}
\STATE $queues\_all[i] := \mathrm{empty}$
\ENDFOR
\STATE $to\_search\_from := \varnothing$
\STATE {$\mathsf{propagate}( \P,queue\_curr,queues\_all, search\_from)$}
\FORALL {$1 \leq i \leq n$}
\STATE $queue\_curr := queues\_all[i]$
\IF {$queue\_current \neq \mathrm{empty}$}
\STATE {$\mathsf{propagate}( \P,queue\_curr, queues\_all, search\_from)$}\label{line:recursive_call}
\ENDIF
\ENDFOR
\STATE $cycle := \epsilon$
\STATE order $search\_from$ decreasingly according to $p.dist$ \label{line:while}
\WHILE {$search\_from \neq \mathrm{empty}$}
\STATE $p := search\_from.\mathsf{top\_and\_pop}()$
\STATE $path\_\mathit{suf} := \mathsf{find\_path}(\P,p,p_f)$
\IF {$path \neq \epsilon$}
\STATE $p_f.dist := p.dist; path\_pref := \epsilon$
\REPEAT
\STATE $path\_pref := (p.pred) \cdot (path\_pref); p:=p.pred$
\UNTIL {$p=p_f$}
\RETURN {$cycle:= (path\_pref) \cdot (path\_suf)$}
\ENDIF
\ENDWHILE \label{line:endwhile}
\RETURN $cycle:=\epsilon$
\end{algorithmic}
\label{alg:cycle}
\end{algorithm}
\normalsize

\begin{algorithm}[h!]
\caption{ \small{$\mathsf{propagate}(\P,queue\_curr, queues\_all,search\_from)$}}
\begin{algorithmic}[1]
\small
\REPEAT
\STATE $p := queue\_curr.\mathsf{front\_and\_pop}()$ \label{line:2}
\IF {$p \not \in visited\_inner$}
\STATE $visited\_inner:=visited\_inner\cup \{p\}$
\FORALL {$p' \in \mathsf{succs}(p)$}
\IF {$p'.dist<p.dist + \W_\P(p,p')$}
\STATE $p'.dist := p.dist + \W_\P(p,p')$; $p'.pred := p$
\IF {$\jana{component}(p') = \jana{(0,0)} \wedge p' \not \in search\_from$ }
\STATE {$search\_from = search\_from \cup \{p'\}$}
\ENDIF
\IF {$\jana{component}(p') = \jana{component}(p)$}
\STATE $queue\_curr.\mathsf{push}(p')$
\ELSIF {$\jana{component}(p') = i$ for some $i \geq 1$}
\STATE $queues\_all[i].\mathsf{push}(p')$
\ENDIF
\ENDIF
\ENDFOR
\ENDIF
\UNTIL {$queue\_curr= \mathrm{empty}$}
\end{algorithmic}
\label{alg:propagate}
\end{algorithm}

\normalsize 

\bigskip

\subsection{Algorithm Summary and Analysis}

The overall solution can be summarized as follows:
\begin{enumerate}
\item Each of the formulas $\phi \in \spec$ is translated into a generalized B\"uchi automaton $\G_\phi$ 
\item A weighted B\"uchi automaton $\B$ is built (see Def.~\ref{def:weighted})
\item A weighted product automaton $\P = \T \otimes \B$ is constructed (see Def.~\ref{def:product}).
\item Alg.~\ref{alg:ndfs} is run on $\P$.
\end{enumerate}

\subsubsection*{Correctness and Correctness}
Based on Lemmas~\ref{lemma:1}--\ref{lemma:reachability} and Propositions~\ref{prop:pf-sf}--\ref{prop:projection}, the soundness and completeness properties of the algorithm are summarized in the following theorem.
\myskipabove
\begin{theorem}[Soundness and completeness]
Given a transition system $\T$, a set of LTL formulas $\spec$ and the reward function $rew$, the suggested algorithm returns \jana{the} solution to Prob.~\ref{prob:reward}.
\end{theorem}


\begin{theorem}
Let $|\T|$ and $|\spec|$ denote the size of the input transition system and the size of the missions specification, respectively. The computational complexity of  Alg.~\ref{alg:ndfs} is in $\mathcal O (|\P| \cdot \log |\P|)$, where $|\P|$ is the size of the product automaton, which is in $\mathcal O (|\T| \cdot 2^{\mathcal O(|\spec|)})$.
\end{theorem}

\myskip
\subsubsection*{Discussion} The translation from an LTL formula $\phi$ into a generalized B\"uchi automaton can be done in  in $2^{\O(|\phi|)}$ time and space. In particular, one of the well-known translation algorithms~\cite{GaOd-cav01} transforms $\phi$ into a generalized B\"uchi automaton with at most $2^{|\phi|}$ states and $|\phi|$ sets in its acceptance condition. If the obtained GBAs for specifications $\phi_1\ldots \phi_n \in \spec$ are all non-blocking, the worst-case size of $\B$ is $2^{(|\spec|)}\cdot(1+|\spec|)$. On the other hand, in case $k$ of the obtained GBAs are blocking, the worst-case size of $\B$ is $2^{(|\spec| + k)}\cdot(|\spec|+k+1)$. Although the size of the resulting GBA is exponential with respect to the size of the input specification, the sizes of the individual formulas are usually small and in many cases, the GBAs are significantly smaller than the worst-case bound. Many optimizations techniques have been also developed among the formal methods literature to reduce the sizes of the GBAs.

The size of the product automaton $\P$ is $|\T| \cdot |\B|$ in the worst case, with at most $|\T| \cdot 2^{(|\spec| + k)}$ in one \jana{component}, where $k$ is the number of blocking GBAs obtained in translation of the formulas from $\spec$. The 
cumulative number of steps made in sorting the set $search\_from$ on line~\ref{line:while} of Alg.~\ref{alg:cycle} is bounded by $\mathcal O (|L_\P|\cdot \log |L_\P|)$, where $|L_\P| = \{p \in Q_\Prod \mid \jana{component}(p) = (0,0)\}$ is the size of the initial \jana{component} of $\P$. 
Altogether, the complexity of Alg.~\ref{alg:ndfs} is in $\mathcal O (|\P| + |L_\P|\cdot \log |L_\P|)$.

In contrast, the "brute-force" approach that tries to find a trace satisfying $\spec_I$, for each $\spec_I \subseteq \spec$ has the worst time complexity characterized as follows. A B\"uchi automaton $\BA_{\spec_I}$ for $\spec_I$ can be constructed with $2^{|\spec_I|} \cdot |\spec_I|$ number of states in the worst case. A nested DFS algorithm is then run on $\P = \T \cdot \BA_{\spec_I}$, reaching complexity $\mathcal O(|\P|)$. Hence, the solution is linear with respect to the size of $|\T| \cdot \sum_{\spec_I \subseteq \spec} 2^{|\spec_I|} \cdot |\spec_I|$.

The benefit of our algorithm (Alg.~\ref{alg:ndfs}) in comparison to the brute-force solution increases with the increasing number of \emph{non-blocking} GBAs obtained from the translation from LTL formulas. Note, that for some LTL formulas, the smallest existing corresponding GBA is non-blocking. In particular many useful specifications, such as $\Event \phi$ (reachability), $\Always \,\Event \, \phi$ (surveillance), $ \Always \, \Event \, \phi_1 \Rightarrow \Always \, \Event \, \phi_2  $ (reactivity), $\Always(\phi_1 \Rightarrow \Event \phi_2)$ (response), or $\Event (\phi_1 \wedge \Event \phi_2)$ (sequencing), where $\phi,\phi_1,\phi_2$ are arbitrary Boolean combinations of atomic propositions, belong to this class.

\section{Rescue Mission Example}
\label{sec:example}

\begin{figure}
\centering
\begin{tabular}{cc}
\subfloat[]{\includegraphics[width=0.4\columnwidth]{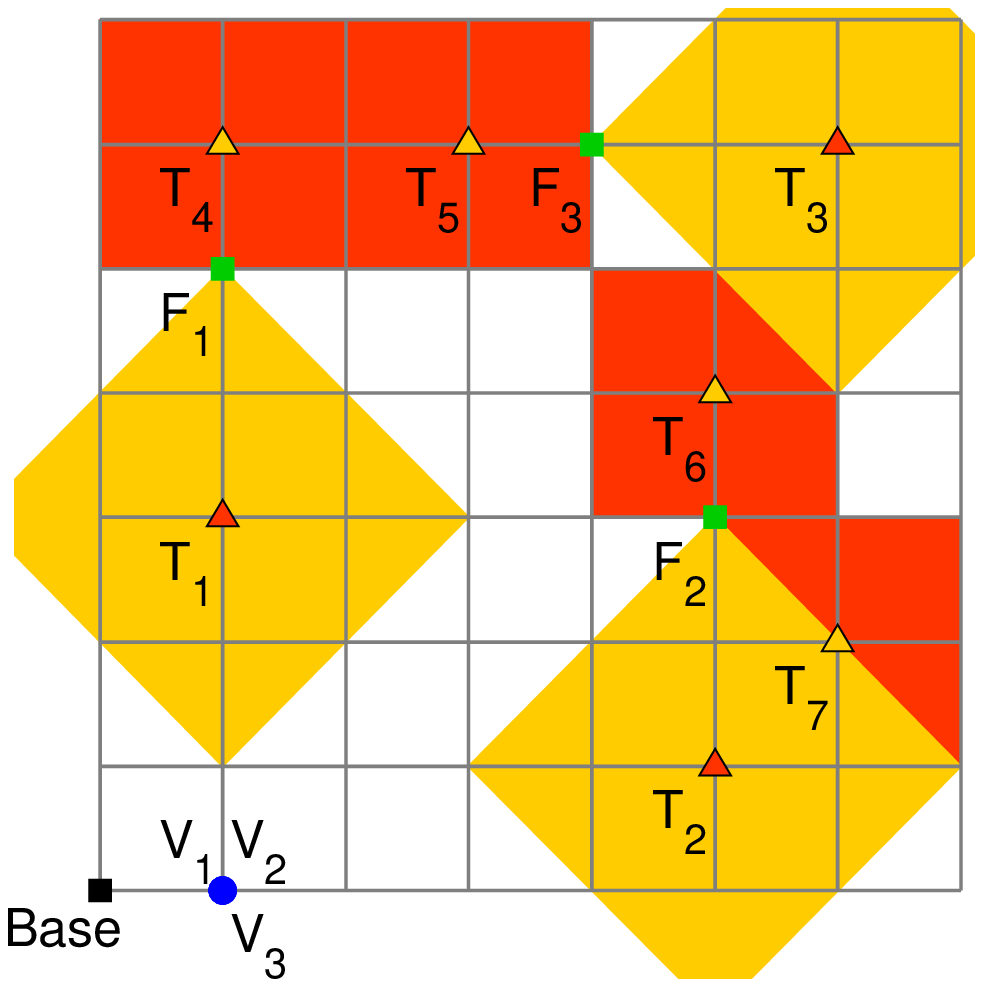}} & 
\subfloat[]{\includegraphics[width=0.4\columnwidth]{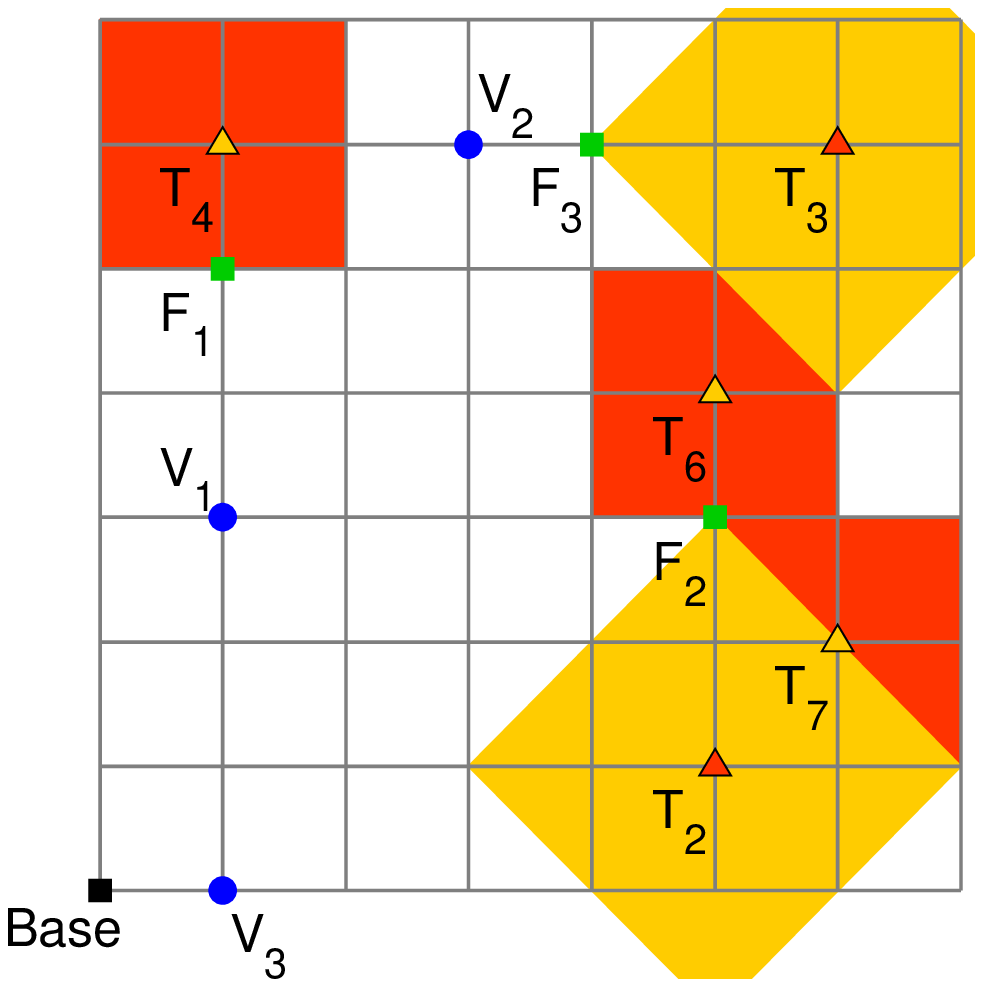}} \\
\subfloat[]{\includegraphics[width=0.4\columnwidth]{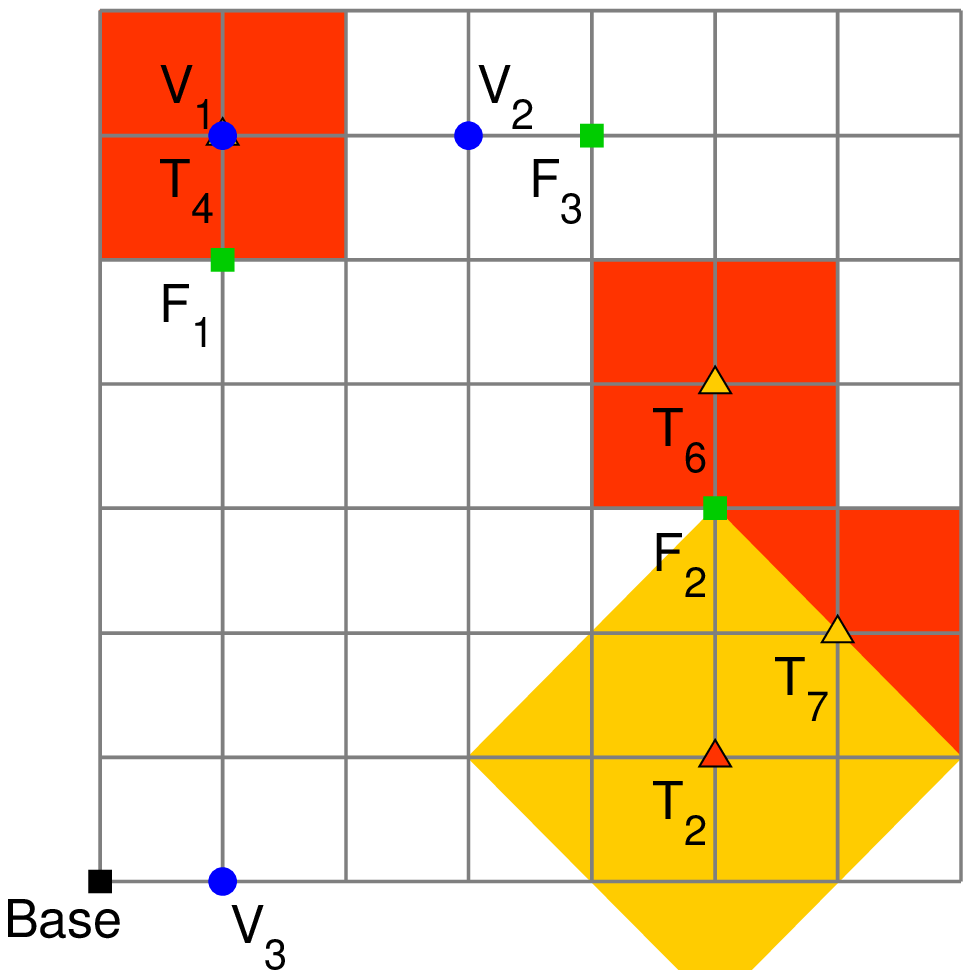}} & 
\subfloat[]{\includegraphics[width=0.4\columnwidth]{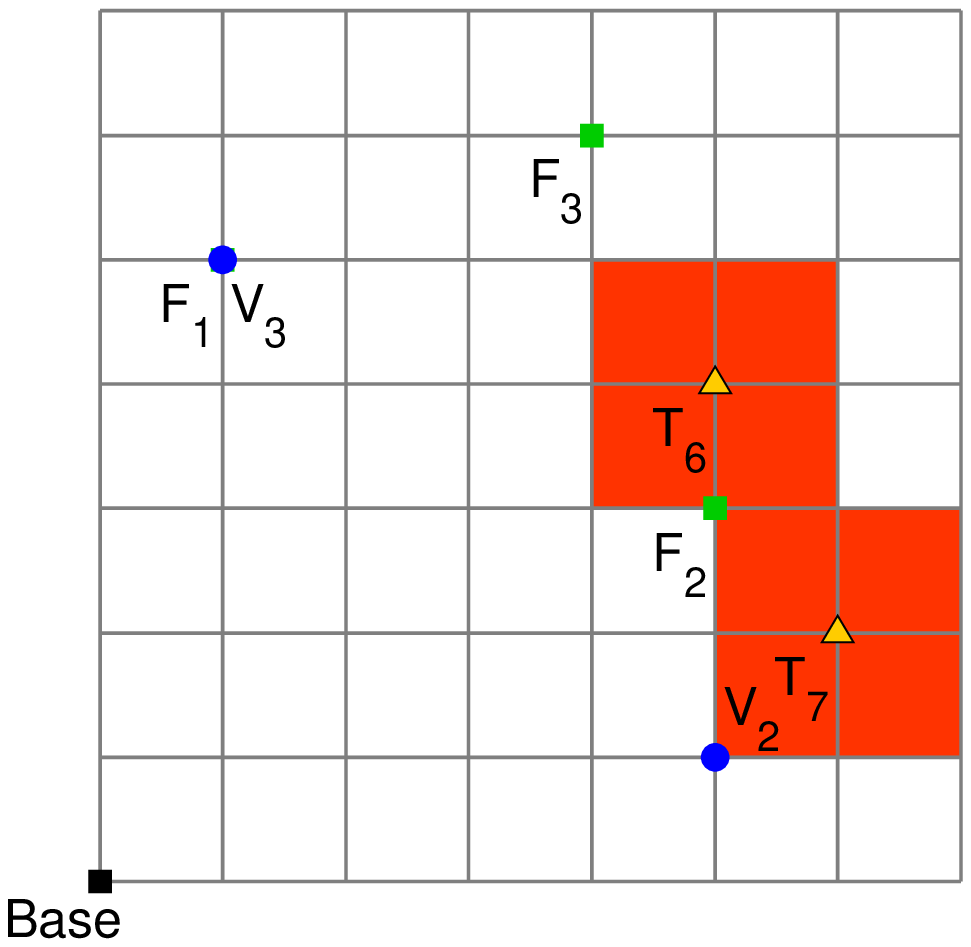}}
\end{tabular}
\caption{An example of a mission featuring conflicting specifications. The captured friendly units $F_1, F_2,$ and $F_3$ are shown as green squares and the enemy units (the targets) $T_1,\ldots, T_7$ are illustrated as triangles. The respective firing ranges of the targets are depicted as yellow and red squares around the targets. The friendly vehicles $V_1,V_2,V_3$ are the blue dots, that can move along the edges of the rectangular grid. A visit of vehicle $V_1$ or $V_2$ into a location with a target is considered an engagement of the target. On the other hand, vehicle $V_1, V_2$, or $V_3$ entering a region within the firing range of a target to which it is vulnerable results in the loss of the vehicle. These rules are captured through irreversible transitions of the underlying state transition system. 
}
\label{fig:rescue}
\end{figure}

Let us consider an example of a complex military rescue mission. Assume that friendly units $F_1, F_2, F_3$ have been captured in an enemy territory. They are guarded by enemy units (called targets) $ T_1, \ldots, T_7$, which need to be engaged before an autonomous vehicle can proceed to pick up the captured friendly units and bring them to the friendly base. A particular configuration is depicted in Fig.~\ref{fig:rescue}.(a). While friendly units $F_1$ and $F_2$ can be rescued by engaging targets $T_1,T_4$, and $T_2,T_6,T_7$, respectively, unit $F_3$ can be rescued by engaging targets $T_3$ and $T_2$. Suppose that we have two unmanned aerial vehicles (UAVs) $V_1$ and $V_2$ and an autonomous ground vehicle $V_3$ under our command, with their capabilities and weaknesses as described below. 
\begin{itemize}
\item $V_1$ can engage $T_1,T_3$, is vulnerable to $T_2,T_5$, and can engage $T_4,T_6,T_7$ at the cost of self-destruction (\ie it can be sacrificed to engage a target $T_4,T_6$, or $T_7$). 
\item $V_2$ can engage $T_2,T_5$, is vulnerable to $T_1,T_3$, and can engage $T_4,T_6,T_7$ at the cost of self-destruction. 
\item $V_3$ can pickup and transport $F_1,F_2,F_3$, but is vulnerable to all active targets. 
\end{itemize}

The mission is to rescue and pickup the friendly units $F_1,F_2$ and $F_3$ and bring them to the base ($Base$). At the same time, the goal is not to loose any of the vehicles $V_1, V_2, V_3$. Let $p_{V_i}^{F_j}$, $p_{V_k}^{Base}$ and $a_{V_\ell}$ denote the atomic propositions ``Vehicle $V_k$ is at the location of the friendly unit $F_j$'', ``Vehicle $V_i$ is at $Base$'', and ``Vehicle $V_\ell$ is active'', respectively. Individual goals are expressed as LTL formulas (see Table~\ref{table:rewards}) and assigned priorities through the reward function. The reward function, among others, specifies that saving the friendly units is more important than not loosing the vehicles $V_1,V_2,V_3$. Note, that because enemy target $T_7$ cannot be destroyed by any of the vehicles at no cost to their integrity, at least one vehicle must be sacrificed to save the friendly units. Although not so obvious, one can also observe that friendly units $F_1$ and $F_2$ cannot both be rescued.

 \renewcommand{\arraystretch}{1.4}
\begin{table}
\centering
\begin{tabular}{ | m{0.45\columnwidth} | m{0.3\columnwidth} | c | } \hline 
\textbf{Mission Specification} & \textbf{LTL Formula:} $\phi$ & $rew(\phi)$ \\ \hline
Pickup $F_i$, and bring it to $Base$, for all $i \in \{1,2,3\}$ & $\Event \: ( \: p_{V_3}^{F_i} \: \wedge \: \Event \: ( \: p_{V_3}^{Base} \: ) \: ),$ for $i \in \{1,2,3\}$ & 10 \\ \hline
Do not pick up $F_3$ before picking up $F_i$, for all $i\in\{1,2\}$ &$\Always \: ( \: p_{V_3}^{F_3} \: \Rightarrow \: \Always \: ( \: \neg \: p_{V_3}^{F_i} \: ) \: ),$ for $i \in \{1,2\}$ & 10 \\ \hline
Do not lose vehicle $V_k$ and bring $V_k$ to $Base$, for all $k\in\{1,2,3\}$ & $\Always \: ( \: \neg \: a_{V_k} \: ) \: \wedge \: \Event \: ( \: p_{V_k}^{Base} \: )$, for $k\in \{1,2,3\}$ & 1 \\ \hline
\end{tabular}
\caption{Mission specification.}
\label{table:rewards}
\end{table}

In order to validate our algorithm, we developed a  \texttt{C++} implementation which takes as an input a deterministic transition system and a list of generalized B\"uchi automata obtained from the LTL formulas with the use of an off-the-shelf tool such as LTL2BA~\cite{ltl2ba}.
The reward gained if the optimal control strategy of the vehicles is applied is \jana{32} units, as expected. Figures \ref{fig:rescue}.(b)--\ref{fig:rescue}.(d) illustrate different stages of the system run. First, vehicles $V_1$ and $V_2$ engage enemy targets $T_1$ and $T_5$, respectively (Fig.~\ref{fig:rescue}.(b)). Then, $V_1$ destroys enemy target $T_3$ before launching a self-destructive attack on $T_4$ (Fig.~\ref{fig:rescue}.(c)). Later, vehicle $V_2$ engages enemy target $T_2$, and vehicle $V_3$ proceeds to pickup $F_1$ and $F_3$, in that order (see Fig.~\ref{fig:rescue}.(d)). Finally, the remaining vehicles return to $Base$. 

\section{Conclusion and Future Work}

In this paper, we have studied the least-violating controller synthesis problem, \ie roughly speaking, to find a trajectory that satisfies the most important pieces of the specification, when the specification can not be satisfied as a whole. We have proposed an algorithm that provides substantial computational savings when compared to a straightforward solution. We have analyzed the proposed algorithm in terms of correctness, completeness and computational complexity. We have also demonstrated the performance of the proposed algorithm on an illustrative example.

There are many directions for future work. In particular, synthesis of {\em optimal} strategies that are least violating, and also synthesis of such strategies to be implemented in {\em dynamic environments} are possible directions for future work.

\label{sec:conclusion}


\bibliographystyle{plain}
\bibliography{tumova.reyes.ea.acc13}

\begin{thebibliography}{10}

\bibitem{asimov}
Isaac Asimov.
\newblock {\em I, Robot}.
\newblock Gnome Press, 1950.

\bibitem{principles}
Christel Baier and Joost-Pieter Katoen.
\newblock {\em Principles of Model Checking}.
\newblock MIT Press, 2008.

\bibitem{modelrepair}
Ezio Bartocci, Radu Grosu, Panagiotis Katsaros, C.~R. Ramakrishnan, and
  Scott~A. Smolka.
\newblock Model repair for probabilistic systems.
\newblock In {\em International Conference on Tools and Algorithms for the
  Construction and Analysis of Systems (TACAS)}, pages 326--340.
  Springer-Verlag, 2011.

\bibitem{BhKaVa-icra10}
Amit Bhatia, Lydia~.E. Kavraki, and Moshe~.Y. Vardi.
\newblock Sampling-based motion planning with temporal goals.
\newblock In {\em Proceedings of the {IEEE} International Conference on
  Robotics and Automation (ICRA)}, pages 2689--2696, 2010.

\bibitem{repair}
Francesco Buccafurri, Thomas Eiter, Georg Gottlob, and Nicola Leone.
\newblock Enhancing model checking in verification by {AI} techniques.
\newblock {\em Artificial Intelligence}, 112(1-2):57 -- 104, 1999.

\bibitem{CiRoScTc-vmcai08}
A.~Cimatti, M.~Roveri, V.~Schuppan, and A.~Tchaltsev.
\newblock Diagnostic information for realizability.
\newblock In {\em Proceedings of the International Conference on Verification,
  Model Checking, and Abstract Interpretation (VMCAI)}, pages 52--67, Berlin,
  Heidelberg, 2008. Springer-Verlag.

\bibitem{CoYa-tac98}
Costas Courcoubetis and Mihalis Yannakakis.
\newblock Markov decision processes and regular events.
\newblock In {\em {IEEE} Transactions on Automatic Control}, 1998.

\bibitem{DaFi-fm11}
Werner Damm and Bernd Finkbeiner.
\newblock Does it pay to extend the perimeter of a world model?
\newblock In {\em Proceedings of the International Symposium on Formal Mehods
  (FM)}, pages 12--26, Berlin, Heidelberg, 2011. Springer-Verlag.

\bibitem{Fa-icra11}
Georgios~E. Fainekos.
\newblock Revising temporal logic specifications for motion planning.
\newblock In {\em Proceedings of the {IEEE} International Conference on
  Robotics and Automation (ICRA)}, 2011.

\bibitem{GaOd-cav01}
Paul Gastin and Denis Oddoux.
\newblock Fast {LTL} to {B}\"uchi automata translation.
\newblock In {\em Proceedings of International Conference on Computer Aided
  Verification (CAV)}, pages 53--65, London, UK, UK, 2001. Springer-Verlag.

\bibitem{ltl2ba}
Paul Gastin and Denis Oddoux.
\newblock {LTL2BA} tool, viewed September 2012.
\newblock URL: http://www.lsv.ens-cachan.fr/~gastin/ltl2ba/.

\bibitem{GePeVaWo-96}
Rob Gerth, Doron Peled, Moshe~Y. Vardi, and Pierre Wolper.
\newblock Simple on-the-fly automatic verification of linear temporal logic.
\newblock In {\em Proceedings of the Fifteenth IFIP WG6.1 International
  Symposium on Protocol Specification, Testing and Verification XV}, pages
  3--18, London, UK, UK, 1996. Chapman \& Hall, Ltd.

\bibitem{Ha-wafr12}
Kris Hauser.
\newblock The minimum constraint removal problem with three robotics
  applications.
\newblock In {\em Proceedings of the International Workshop on the Algorithmic
  Foundations of Robotics (WAFR)}, 2012.

\bibitem{spin}
Gerard~J. Holzmann.
\newblock {\em The Spin Model Checker: Primer and Reference Manual}.
\newblock Addison-Wesley Professional, 2003.

\bibitem{KaFr-cdc09}
Sertac Karaman and Emilio Frazzoli.
\newblock Sampling-based motion planning with deterministic $\mu$-calculus
  specifications.
\newblock In {\em Proceedings of the {IEEE} Conference on Decision and Control
  (CDC)}, pages 2222--2229, 2009.

\bibitem{KaFr-acc12}
Sertac Karaman and Emilio Frazzoli.
\newblock Sampling-based optimal motion planning with deterministic
  $\mu$-calculus specifications.
\newblock In {\em Proceedings of the American Control Conference (ACC)}, 2012.

\bibitem{KiFaSa-icra12}
Kangjin Kim, Georgios Fainekos, and Sriram Sankaranarayanan.
\newblock On the revision problem of specification automata.
\newblock In {\em Proceedings of the {IEEE} International Conference on
  Robotics and Automation (ICRA)}, 2012.

\bibitem{KlBe-tac08}
Marius Kloetzer and Calin Belta.
\newblock A fully automated framework for control of linear systems from
  temporal logic specifications.
\newblock {\em {IEEE} Transactions on Automatic Control}, 53(1):287 --297,
  2008.

\bibitem{KrFaPa-tac09}
Hadas Kress-Gazit, Georgios~E. Fainekos, and George~J. Pappas.
\newblock Temporal-logic-based reactive mission and motion planning.
\newblock {\em {IEEE} Transactions on Automatic Control}, 25(6):1370--1381,
  2009.

\bibitem{LaWaAnBe-icra10}
Morteza Lahijanian, Joe Wasniewski, Sean~.B. Andersson, and Calin Belta.
\newblock Motion planning and control from temporal logic specifications with
  probabilistic satisfaction guarantees.
\newblock In {\em Proceedings of the {IEEE} International Conference on
  Robotics and Automation (ICRA)}, pages 3227 --3232, 2010.

\bibitem{RaKr-cav11}
Vasumathi Raman and Hadas Kress-Gazit.
\newblock Analyzing unsynthesizable specifications for high-level robot
  behavior using ltlmop.
\newblock In {\em Proceedings of International Conference on Computer Aided
  Verification (CAV)}, pages 663--668, 2011.

\bibitem{RaKr-icra12}
Vasumathi Raman and Hadas Kress-Gazit.
\newblock Automated feedback for unachievable high-level robot behaviors.
\newblock In {\em Proceedings of the {IEEE} International Conference on
  Robotics and Automation (ICRA)}, pages 5156--5162, 2012.

\bibitem{SmTuBeRu-ijrr11}
Stephen.~L. Smith, Jana Tumova, Calin Belta, and Daniela Rus.
\newblock Optimal path planning for surveillance with temporal logic
  {C}onstraints.
\newblock {\em International Journal of Robotics Research}, 30(14):1695--1708,
  2011.

\bibitem{WoToMu-cdc09}
Tichakorn Wongpiromsarn, Ufuk Topcu, and Richard~M. Murray.
\newblock Receding horizon temporal logic planning for dynamical systems.
\newblock In {\em Proceedings of the {IEEE} Conference on Decision and Control
  and the Chinese Control Conference (CDC/CCC)}, pages 5997 --6004, 2009.

\end{thebibliography}

\end{document}